%% file: main.tex
\documentclass{article}
\usepackage[utf8]{inputenc}
\usepackage{amssymb}
\usepackage{times}  % DO NOT CHANGE THIS
\usepackage{helvet}  % DO NOT CHANGE THIS
\usepackage{courier}  % DO NOT CHANGE THIS
\usepackage[hyphens]{url}  % DO NOT CHANGE THIS
\usepackage{graphicx} % DO NOT CHANGE THIS
\usepackage{caption} % DO NOT CHANGE THIS AND DO NOT ADD ANY OPTIONS TO IT
\usepackage{algorithm}
\usepackage{algorithmic}
\usepackage{times}
\usepackage{epsfig}
\usepackage{graphicx}
\usepackage{xspace}
\usepackage{amsmath}
\usepackage{wrapfig,lipsum,booktabs}

\usepackage{verbatim}
\usepackage{enumitem}
\usepackage{makecell}
\usepackage{stfloats}
\usepackage{authblk}
\usepackage{apacite}
\usepackage[normalem]{ulem}

\usepackage[a4paper,
            bindingoffset=0.2in,
            left=1.25in,
            right=1.25in,
            top=1.25in,
            bottom=1.25in,
            footskip=.25in]{geometry}

\usepackage{blindtext}

\newcommand{\datasetname}[0] {VISUELLE}
\newcommand{\approachname}[0] {GTM-Transformer\xspace}

\title{Well Googled is Half Done: Multimodal Forecasting of New Fashion Product Sales with Image-based Google Trends}

\author[1]{Geri Skenderi}
\author[2]{Christian Joppi}
\author[2]{Matteo Denitto}
\author[1,2]{Marco Cristani}

\affil[1]{Department of Computer Science, University of Verona, Italy}
\affil[ ]{\{name.surname\}@univr.it}
\affil[2]{Humatics Srl, Verona, Italy}
\affil[ ]{\{name.surname\}@sys-datgroup.com}

\begin{document}

\maketitle

\abstract{New fashion product sales forecasting is a challenging problem that involves many business dynamics and cannot be solved by classical forecasting approaches. In this paper, we investigate the effectiveness of systematically probing exogenous knowledge in the form of Google Trends time series and combining it with multi-modal information related to a brand-new fashion item, in order to effectively forecast its sales despite the lack of past data. In particular, we propose a neural network-based approach, where an encoder learns a representation of the exogenous time series, while the decoder forecasts the sales based on the Google Trends encoding and the available visual and metadata information. Our model works in a non-autoregressive manner, avoiding the compounding effect of large first-step errors. As a second contribution, we present \datasetname{}, a publicly available dataset for the task of new fashion product sales forecasting, containing multimodal information for 5577 real, new products sold between 2016-2019 from Nunalie, an Italian fast-fashion company. The dataset is equipped with images of products, metadata, related sales, and associated Google Trends. We use \datasetname{} to compare our approach against state-of-the-art alternatives and several baselines, showing that our neural network-based approach is the most accurate in terms of both percentage and absolute error. It is worth noting that the addition of exogenous knowledge boosts the forecasting accuracy by 1.5\% in terms of Weighted Absolute Percentage Error (WAPE), revealing the importance of exploiting informative external information. The code and dataset are both available at \url{https://github.com/HumaticsLAB/GTM-Transformer}.}

\section{Introduction}\label{sec:intro}
\input{src/intro}

\section{Related Work}\label{sec:related}
\input{src/related}

\section{The VISUELLE dataset}\label{sec:dataset}
\input{src/dataset}

\section{Method}\label{sec:method}
\input{src/method}

\section{Experiments}\label{sec:exp}
\input{src/exp}

\section{Conclusion}\label{sec:conclusion}
\input{src/conc}

\bibliographystyle{apacite}
\bibliography{bibi.bib}
\end{document}

%% file: src/intro.tex
Sales forecasting is one of the earliest and most demanded forecasting applications~\cite{intelligentFashionBook, KashiSalesSurvey}. Driven by economic and financial reasons, the ability to anticipate the needs and behavior of customers can make a big difference for commercial activity, especially when large volumes of goods need to be managed. While the forecasting of time series with a known historical past has been analysed extensively ~\cite{hyndman_athanasopoulos_2021,benitez_salesForecastingReview_2021}, very little attention has been paid to a much more practical and challenging scenario: the forecasting of new products, which the market has not seen before. In many cases, such forecasts are made in a judgmental manner~\cite{hyndman_athanasopoulos_2021} by experts that essentially take into consideration the characteristics of the newly designed product along with information on what is trending right now in the market to make an educated guess.

In this paper, we propose a non-autoregressive Transformer \cite{vaswani2017attention} model dubbed \approachname{}, which tries to mimic this behavior by modeling the sales of new products based on information coming from several domains (modalities): the product image; textual descriptors of category, color and fabric; temporal information in the planned release date; and exogenous information on the trending tendencies of the textual descriptors in the form of Google Trends. This last component is a crucial part of \approachname{}, since it introduces external information on item popularity into the reasoning. Intuitively, it models what people are interested in and proves important for forecasting performance.

While it has been already shown that Google Trends can be used to predict diverse types of time series (from real estate sales to inflation expectations)~\cite{wu20153,bulut2018google,hand2012searching,hamid2015forecasting,guzman2011internet,bangwayo2015can}, their adoption to clothing sales forecasting has only been suggested in~\cite{silva2019googling} but never tried in practice, especially in a new product forecasting setting. Technically, we demonstrate that Google Trends are valuable when encoded appropriately. Thanks to the Cross-Attention weights of our model, we find that the most useful information is systematically located around the end of the previous year's same fashion season, i.e., seven to ten months before the product is planned for exposure.

As a second contribution, we present \datasetname{}: the first public dataset for new fashion product sales forecasting. \datasetname{} is a repository build upon the data of a real fast fashion company, Nunalie\footnote{\url{http://www.nunalie.it .}} and is composed of 5577 new products and about 45M sales related to fashion seasons from 2016-2019. Each product in \datasetname{} is equipped with multimodal information: an image, textual metadata, sales after the first release date, and three related Google Trends describing \emph{category}, \emph{color} and \emph{fabric} popularity.  We use \datasetname{} to compare \approachname{} with the few and recent state-of-the-art alternatives in the new product sales forecasting literature, obtaining the best performance on several forecasting metrics. We also show that the model can be enriched with attributes which are automatically inferred from the image, considering the widely-used Fashion IQ attributes~\cite{wu2020fashion}, ameliorating the final performance. 

The rest of the paper is organized as follows: the ensuing Sec. will provide a general overview of the literature around forecasting in fashion and new product sales forecasting. In Sec. \ref{sec:dataset}, an overview of the \datasetname{} dataset is given, showing the available information and how the dataset can be used for further research on this topic. Sec. \ref{sec:method} explains the methodological background and details behind \approachname{}. In Sec. \ref{sec:exp}, the experiments are thoroughly explained and finally, conclusions are drawn in Sec. \ref{sec:conclusion}. 

%% file: src/related.tex
\subsection{New product sales forecasting}
Tackling the new product sales forecasting with machine learning tools has very few precedent cases~\cite{ekambaram_attention_2020,singh_fashion_2019}. The intuition followed in general is that new products will sell comparably to similar, older products; consequently, these models should be able to understand similarities among new and older products. 

In~\cite{singh_fashion_2019}, a variety of boosting algorithms (XGBoost, Random Forest) and Neural Networks (MLP, LSTM) are taken into account, fed with textual attributes related to category and colors, and merchandising factors such as discounts or promotions. Notably, they do not make use of image features or exogenous information. The most related work with ours is~\cite{ekambaram_attention_2020}, where the authors use an autoregressive RNN model that takes past sales, auxiliary signals like the release date and discounts, textual embeddings of product attributes, and the product image as input. The model uses soft-attention to understand which of the modalities is the most important to the sales. The model then embeds and combines all these attended features into a feature vector which is fed to a GRU \cite{cho2014learning} decoder and used to forecast the item sales. In contrast to our work, ~\cite{ekambaram_attention_2020} do not make use of a "true exogenous" signal such as the Google Trends, the model is based on internal information available in the data. Additionally, the autoregressive nature of RNNs creates prediction curves which have a very common shape across products. Unfortunately the dataset and the code is proprietary and was not released. Recent work has explored the use of exogenous signals in fashion forecasting, with \cite{joppi2022pop} showing that it is possible to build informative popularity signals that are helpful for forecasting. Our works adds to the literature by using available popularity signals from the web and a Transformer-based model to predict the sales in one shot, without autoregression.

\subsection{Datasets for fashion forecasting}
Publicly available datasets for fashion forecasting take into account diverse applications, dissimilar from new product forecasting. The "Clothing, shoes and jewelry" dataset has been used in~\cite{ni-etal-2019-justifying,al-halah_fashion_2020} to forecast fashion styles, that is aggregates of products of multiple brands, in terms of popularity on Instagram. In our case the problem is different, since we are focusing on \emph{single} products and not on groups of products, so we have definitely fewer data to reason on. In addition, we are considering \emph{genuine sales data}, and not popularity trends. This makes our research more impactful on an industrial level. The Fashion Instagram Trends\cite{ma_knowledge_2020} adds geographical information to forecast trends in specific places.
%-------------------------------------------------------------------------

%% file: src/dataset.tex
\begin{figure*}[t]
        \includegraphics[width=\linewidth]{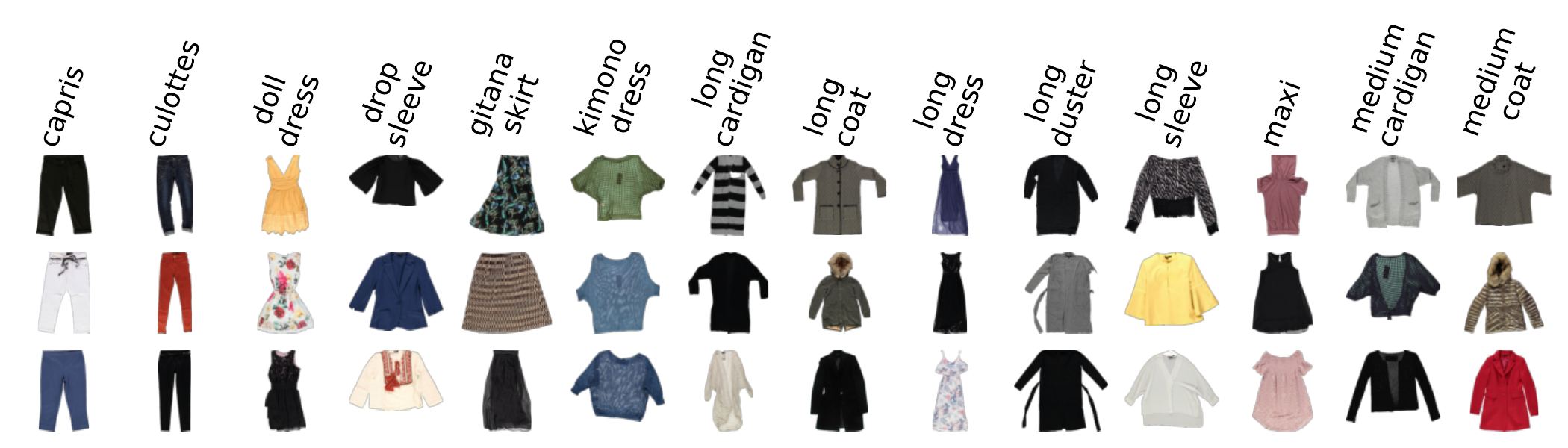}
        \includegraphics[width=\linewidth]{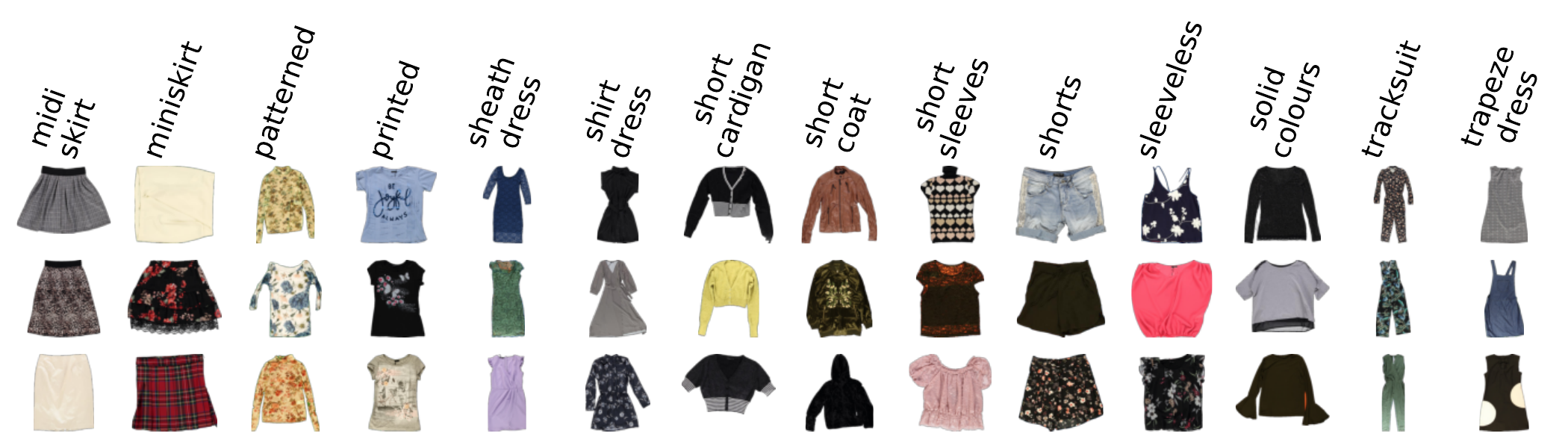}
        \caption{Sample images representing various product categories within the \datasetname{} dataset.}
        \label{fig:mosaic}
    \end{figure*}

\datasetname{} describes the sales between October 2016 and December 2019 of 5577 products in 100 shops of Nunalie, an Italian fast-fashion company funded in 2003. For each product, multimodal information is available, which will be detailed in the following subsections.

\subsection{Image data} Each product is associated with an RGB image, of resolution which varies from 256 to 1193 (width) and from 256 to 1172 (height) with median values 575 (w) 722 (h) . Images have been captured in a controlled environment, in order to avoid color inaccuracies and potential biases in the predictions~\cite{nitse2004impact}. Each image portrays the clothing item on a white background, with no person wearing it. Additionally, a binary foreground mask is provided.

\begin{figure*}[t]
        \includegraphics[width=\linewidth]{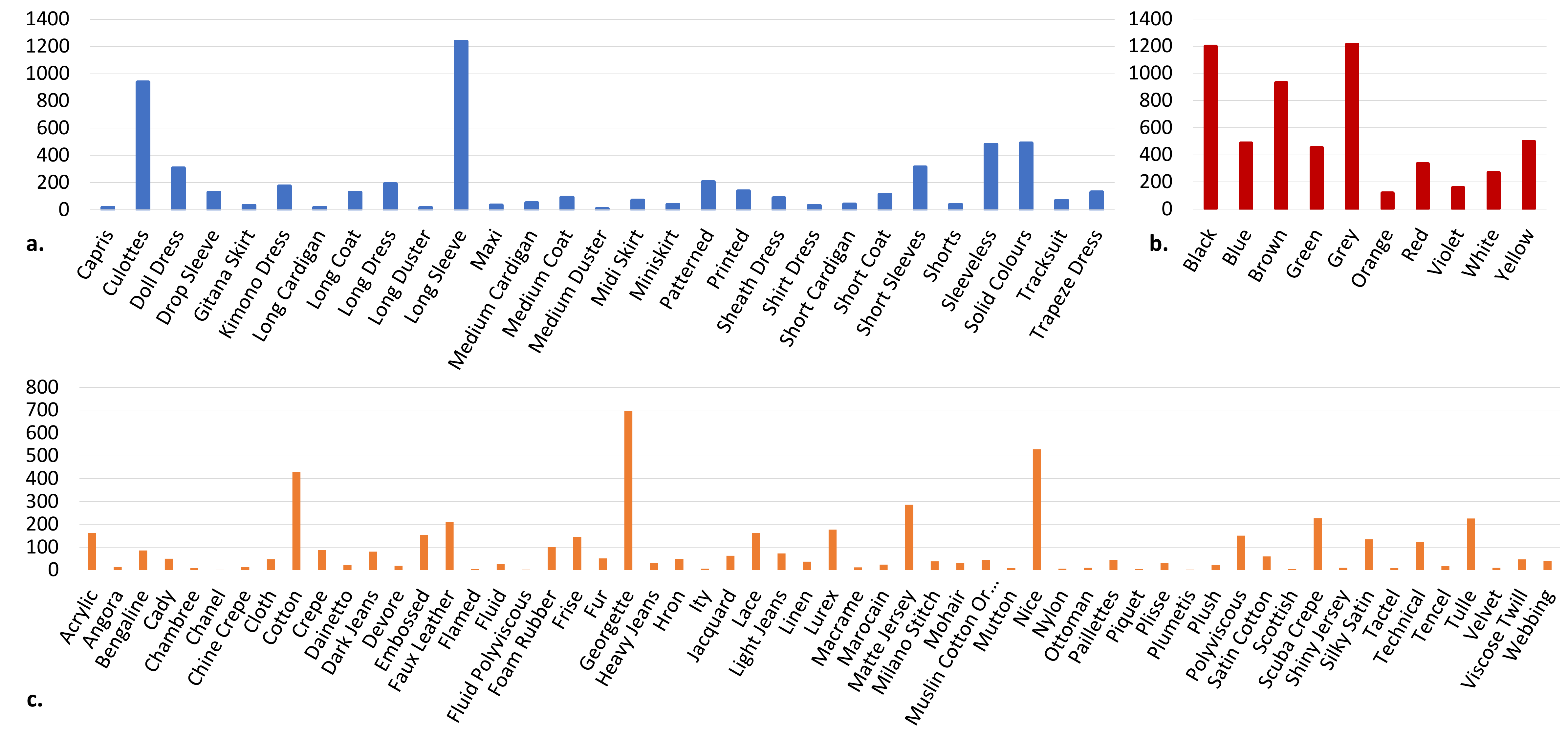}
        \caption{Cardinalities of the dataset for clothing categories (a), color (b) and fabric (c).}
        \label{fig:cardinalities}
\end{figure*}

\begin{figure}[t]
    \centering
    \includegraphics[width=.5\columnwidth]{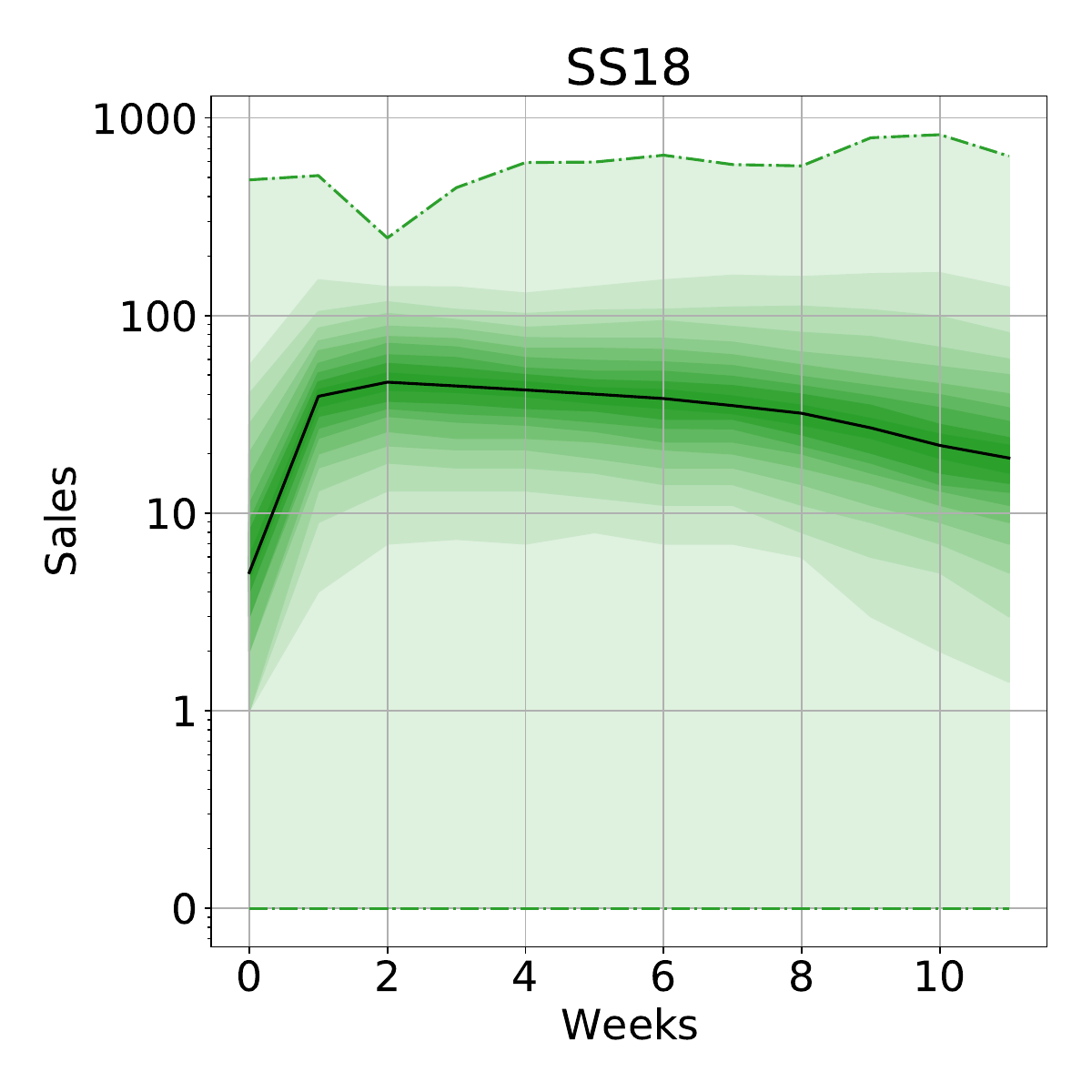}\hfill
    \includegraphics[width=.5\columnwidth]{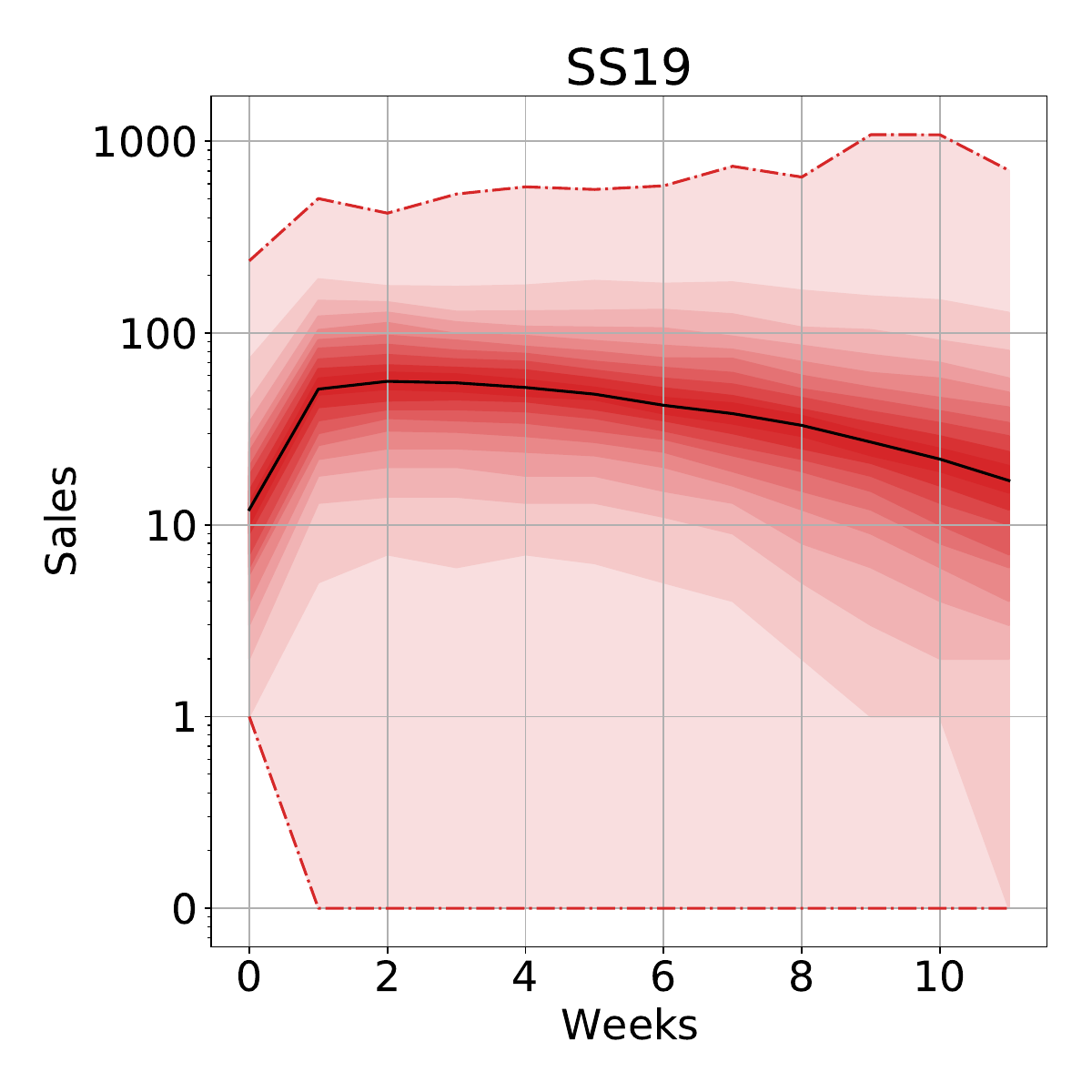}
     \caption{25th-percentile density plots of the SS18 and SS19 seasons.}
    \label{fig:sales}
\end{figure}
\iffalse
\begin{figure}[t]
        \includegraphics[width=\columnwidth]{figures/ss18_19}\hfill
        \includegraphics[width=\columnwidth]{figures/long_sl}\hfill
        \includegraphics[width=\columnwidth]{figures/kimono_dr}\hfill
     \caption{Histograms of the first 4 weeks of sales for seasons SS18 (above) and SS19(below): a) all the categories are shown; b) the most populated category: long sleeves; c) on of the least populated category: kimono skirt.}
    \label{fig:histo}
\end{figure}
\fi

\subsection{Text data} Each product has multiple associated tags, which have been extracted with diverse procedures detailed in the following, and carefully validated by the Nunalie team. The first tag is the \emph{category}, taken from a vocabulary of 27 elements, visualized in Fig.~\ref{fig:cardinalities}a; the cardinality of the products shows large variability among categories, because naturally some of them (e.g. long sleeves) are popular year-round. This fact demonstrates a particular and challenging aspect of \datasetname{}, which renders the learning of sales dynamics over different categories non-trivial. The \emph{color} tag represents the most dominant color, and is extracted from the images with a proprietary pixel clustering algorithm, keeping the color with the most belonging pixels, and validated for each product by two human operators that must agree on it. The final vocabulary is made of 10 elements. The cardinality per color is reported in Fig.~\ref{fig:cardinalities}b. The \emph{fabric} tag describes the material from which clothes are made, and comes directly from the technical sheets of the fashion items. This tag comes from a vocabulary of 58 elements, visualized in Fig.~\ref{fig:cardinalities}c; A product is sold during a particular season, and within a season, released on the market at a precise day. This \emph{temporal information} is recorded as a text string. Holidays and sales periods are supplementary information which we plan to deliver for a second version of the dataset. 

\subsection{Sales data}
The sales time series have a weekly frequency and contain 12 observations each, which corresponds to the permanence of an item in the shops during a fashion season (Autumn-Winter, AW and Spring-Summer, SS). Fig.~\ref{fig:sales} contains a density plot of the sales of all the products, merging together different categories, across corresponding seasons (SS18 and SS19 were used for clarity). This is useful to show that there are general ``mean curves'' where the sales peak occurs after a week and that as the weeks go by, the sales are characterized by a higher variability. An increase of the sales during the years is visible, showing that the company seems to perform well. Notably, from the release moment until 6 weeks, no external action is done by the company owners (discounts, pre/sales, additional supplying) and they had never sold out products, so we can state that the signal variability is given by the product attractiveness.

\begin{figure*}[h]
    \centering
    \includegraphics[width=.3\linewidth]{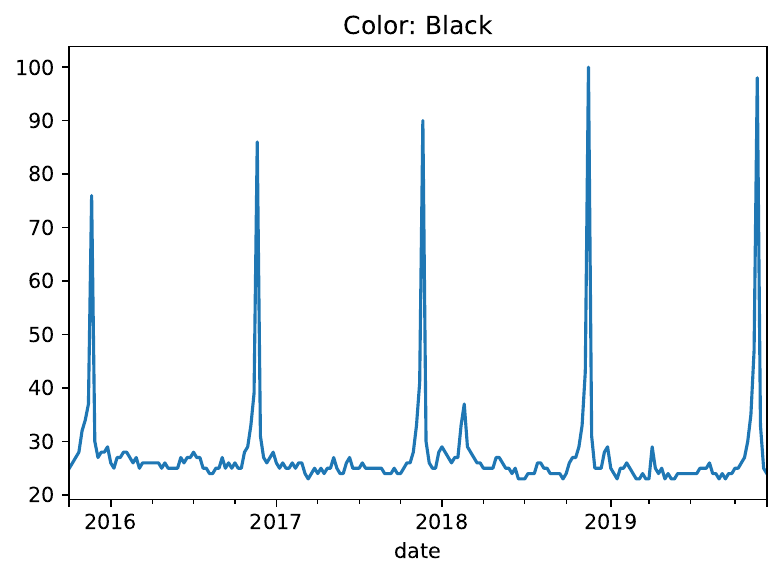}\hfill
    \includegraphics[width=.3\linewidth]{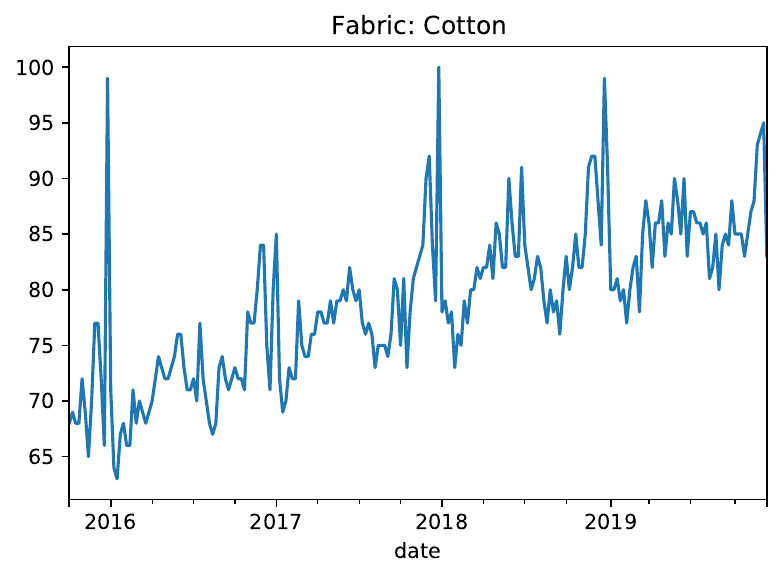}\hfill
    \includegraphics[width=.3\linewidth]{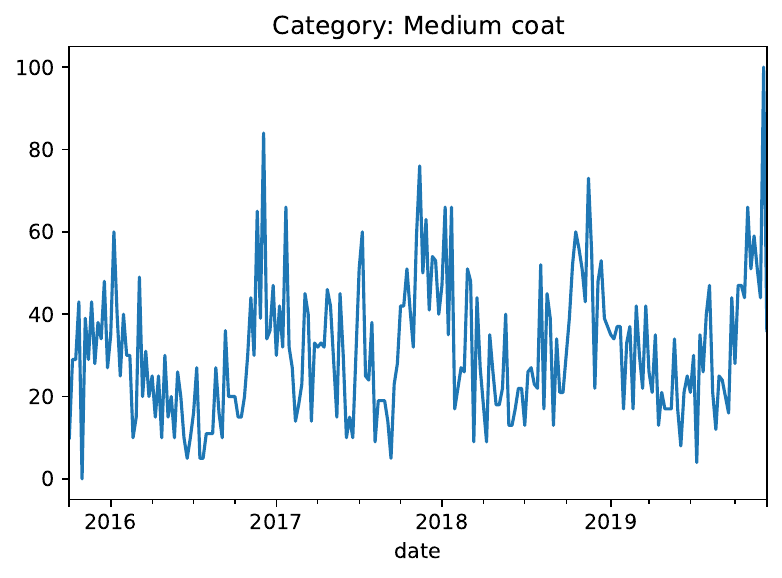}\hfill
    \caption{Examples of Google Trends time-series spanning multiple years.}
    \label{fig:structuredTrend}
\end{figure*}

\subsection{Google Trends data}\label{sec:gtrend_collection}
Extracting Google Trends to discover the popularity of textual term\textbf{s} describing visual data poses a paradox: the more specific the text, the less informative the signal (due to sparsity), and vice-versa. In \datasetname{} we collect, for each product, three Google trends time-series by querying the API using each of the product's three associated textual attributes: ${color, category, fabric}$. The trends are downloaded starting from the release date and going back 52 weeks, essentially anticipating the release of each single item by one year. Each signal represents a percentage, reaching 1 (100\%) in the moment in time when that particular attribute had the maximum search volume on Google, for the duration of the search interval. 

Fig.\ref{fig:structuredTrend} contains examples of Google Trends in the interval 2016-2019. As visible, the nature of these signals is highly variable, spanning from highly structured to more noisy. To make the Google trends signal more reliable, we follow the ``multiple sampling'' strategy discussed in~\cite{medeiros2021proper}. Google normalizes the search results of a query by the total searches of the location and time range chosen by the user. Then, the resulting numbers are scaled from 0 to 100, in order to represent the relative popularity. The problem is of course, because of the high amount of search queries that Google processes each day, the query results are always a sub-sample of the "true" ones and this sample may not always be the same. To avoid this occurrence, we download each Google Trend 10 times and use the mean to create a more representative signal.
%-------------------------------------------------------------------------

%% file: src/method.tex
\subsection{Background}\label{sec:attention}
Given a product $x$, we refer to its \emph{product sales} time series as $S(x,t)$ where $t$ refers to the $t$-th week of market delivery, with $x=1,...,N$ and $t=1,...,T$. In this work, we assume that each product $x$ is associated with an image $i$, a set of textual tags $T$ and a planned release date $d$, as in the \datasetname{} dataset. Finally, exogenous series in the form of Google Trends are assumed to be available or collected as detailed in \ref{sec:gtrend_collection}. The goal is to combine all of this information in an efficient and expressive way, such that we can forecast $S(x,t)$ as accurately as possible. We would like to stress again that in the new fashion product sales forecasting scenario, we cannot have direct access to $S(x,t)$, as the product $x$ is new and therefore does not have past sales.

\begin{figure}[t]
    \centering
    \includegraphics[width=0.9\linewidth]{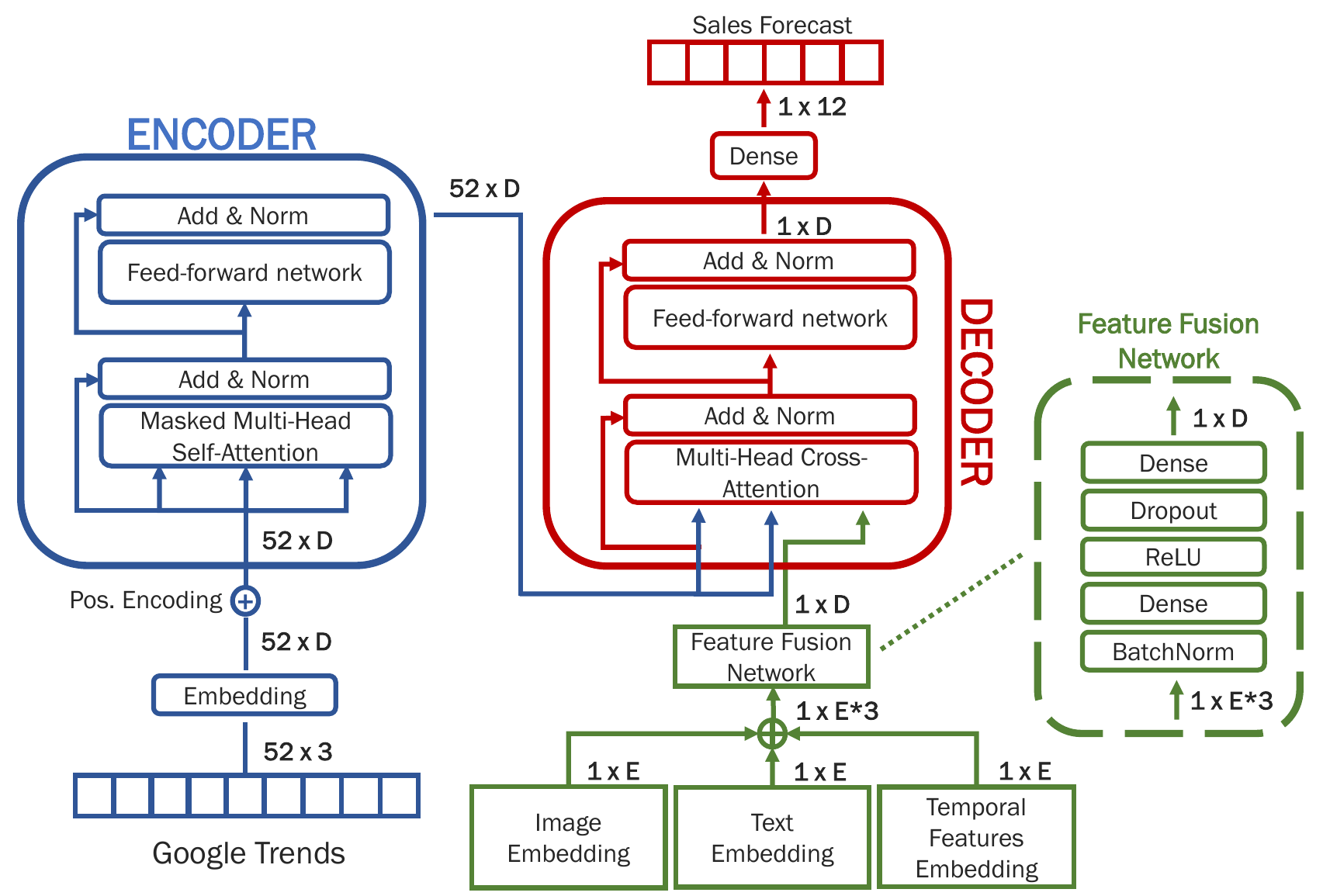}
    \caption{\approachname{} architecture. The encoder processes the exogenous Google Trends series and learns a representative embedding thanks to the self-attention mechanism. The decoder takes as input a multimodal embedding created from the Feature Fusion Network and then relies on a cross-attention mechanism to understand the implications of the Google Trend series on the multimodal embedding for the forecasting task. The output of the transformer model is then passed through a fully connected layer, to \textit{generate} the sales forecast.}
    \label{fig:model_architecture}
\end{figure}

The structure of the proposed model is depicted in Fig.~\ref{fig:model_architecture}: \approachname{} is based on the Transformer model~\cite{vaswani2017attention}, but it deviates from the canonical form in two ways. Firstly, the decoder contains no self-attention block, since the multimodal embedding that acts an input is a dense vector. The use of a non-autoregressive variant~\cite{gu2017non} is motivated by two reasons: i) to avoid the compounding of errors caused by wrong initial predictions; ii) to generate the forecasted time series in one go, without any recurrence mechanism, allowing for faster training and inference. In particular, \approachname{} learns different representations for each input type and then projects such representations in a shared latent space to non-autoregressively forecast the sales. 

Before proceeding with the explanation of the model and its different components, we give a brief explanation of the attention mechanism, since it is the driving force behind the Transformer model~\cite{vaswani2017attention} and most state-of-the-art neural sequence processing models. An attention function can be described in simple words as mapping a query and a set of key-value pairs to an output, where the query, keys, values, and final output are all vectors. The result is calculated as a weighted sum of the values, where the weight of each value is determined by the query's compatibility function with the corresponding key. These terms may sound confusing, but they are actually simple ideas coming from information retrieval systems. For instance, when we search for a query in a search engine, the engine will try to compare the \textit{search query} to a list of \textit{keys} (typically metadata) linked to potential matches in their database, and then return the \textit{values} that give the best matches. To \emph{learn} these query-key compatibilities, also known as attention weights, the Scaled Dot-Product Attention mechanism~\cite{vaswani2017attention} projects the input in a common vector space and then relies on the dot-product to compute the similarity between vectors. The final output is actually a weighted average of the values according to the query-key compatibilities, because the learned attention weights are normalized to sum to one, as described by the following equations:
\begin{align}
     Q & = {W_q}X + B_q, \label{eq:1} \\
     K & = {W_k}X + B_k, \label{eq:2} \\
     V & = {W_v}X + B_v, \label{eq:3} \\
     \alpha & = \frac{(QK^T)}{d_k}, \label{eq:4} \\
     \hat{y} & = softmax(\alpha)V \label{eq:5}
\end{align}
where $W$ and $B$ are learnable weight and bias matrices for each transformation respectively, $d_k$ is the dimensionality of the key($K$) latent space, and the $softmax$ refers to the differentiable approximation of the $argmax$ function, which outputs a categorical probability distribution. By looking at equations \ref{eq:4} and \ref{eq:5}, it becomes clear why the name of this method is "Scaled Dot-Product Attention". This powerful learning technique needs several training tricks to work well with sequences in practice, such as: Positional Encoding; Masking; Learning Rate warmp-up. For a detailed explanation of the Transformer model and its inner workings, we refer the reader to~\cite{vaswani2017attention}.  

\subsection{\approachname{}}
Our architecture is an encoder-decoder sequence model \cite{sutskever2014} that is composed of several blocks, which interact with each other to create a model that is able to accurately forecast $S(x,t)$. To do this we rely mainly on dense vector representations of the different modalities and the Scaled Dot-Product Attention mechanism. The main idea behind the model is to create a mixture between a retrieval system and a forecasting system, where the model first learns to generate an embedding for each product, and later perform a forecast. During the learning procedure, the model is able to learn and then reason on sales associations between similar and dissimilar products. In simpler words, we aim to build an architecture that is capable of learning (in a non-linear way) how to make judgemental forecasts~\cite{FPAP2} of new fashion products. In order to do this, our model relies on its ability to retrieve similar items and understand their sales, as well as the popularity of the item tags that comes from the Google Trends.  The building blocks of \approachname{} and their interactions are described below and depicted in Fig.~\ref{fig:model_architecture}.

\subsubsection{Creating a multimodal product embedding}
The first part of our architecture consists of a model that can effectively fuse the information coming from the different modalities. In order to this, we makes use of separate embedding modules for each modality, which create dense vector representations of the respective data in a shared latent space $R^E$. Afterwards, a Feature Fusion Network takes as input these vector representations and generates a unique, multimodal dense representation for the product. 

\textbf{The image embedding module} relies on a ResNet-50 model~\cite{he2015deep} pre-trained on ImageNet~\cite{imagenet} to extract 2D convolutional features $\phi_{resnet} \in \\R^{CxWxH}$, where $C=2048$ is the number of final feature channels, $W$ is the image width and $H$ the image height. Adaptive average pooling \cite{Liu_2018_CVPR} with a square kernel of size 1 is applied, followed by a Fully Connected Layer, which generates the final image representation using the linear transformation: 
\begin{equation}\label{eq:6}
    \phi_i = W_i \phi_{resnet}  + B_i,\: \phi_i \in R^E.
\end{equation}

\textbf{The text embedding module} relies on the BERT model~\cite{devlin2019bert} pre-trained on a large corpus comprising the Toronto Book Corpus and Wikipedia. This module takes as input the same set of textual queries used to find the Google Trend (Sec. \ref{sec:gtrend_collection}), i.e ${color,category, fabric}$ and produces an embedding $\phi_{bert} \in R^{768}$ of the words. Exploiting a pre-trained language model offers an additional advantage, which is the ability to generate a representation for any textual tag, even those that it might have never seen before. Because these language models are trained on extremely large corpora, they also add additional context that comes the typical uses of the textual tag in a sentence. The module averages the embeddings for each attribute and then uses a Fully Connected Layer to create the final textual representation: 
\begin{equation}\label{eq}
    \phi_t = W_t \phi_{bert}  + B_t,\: \phi_t \in R^E.
\end{equation}

\textbf{The temporal features embedding module}, is an MLP (Multi Layer Perceptron) that creates a set of embeddings $\{\phi_d, \phi_w, \phi_m, \phi_y\}$ that contains a projection of each temporal feature available for each product, extracted from the planned release date: the day of the week, the week of the year, the month and the year. Afterwards, these embeddings are concatenated and merged together through a dense layer, creating a final representation of all these temporal features:
\begin{equation}\label{eq}
    \phi_r = W_r [\phi_d ; \phi_w; \phi_m; \phi_y ]  + B_r,\: \phi_r \in R^E.
\end{equation}
 where $[;]$ is the concatenation operation.

\textbf{The feature fusion network} is another MLP that merges the separate modalities together by using a cascade of fully-connected layers and a non-linear activation. This allows the model to generate an expressive and dense final product embedding in a higher dimensional space $R^D,\: D > E$, which also contains non-linear interactions between the different modalities:
\begin{align}
     \psi_f & = f(\phi_i, \phi_t, \phi_r),\: \psi_f \in R^D \label{eq:9}\\
     f(\phi_i, \phi_t, \phi_r) & = W_{l2}ReLU(W_{l1}[\phi_i; \phi_t; \phi_r]) + B_{l2} \label{eq:10}
\end{align}
where $ReLU$ represents the Rectified Linear Unit activation function \cite{agarap_relu}.

\subsubsection{Forecasting}
\approachname{} first produces a self-attended representation of the Google Trend time series and then merges that information along with the previously explained multimodal embedding in order to forecast the future sales of a new product.

\textbf{The transformer encoder} takes as input a multivariate Google Trends time series (one for each textual attribute of the product). The series are projected in the same latent space $R^D$ and then are enriched with a positional encoding. This signal is then processed by the standard encoder block of \cite{vaswani2017attention}, by applying Scaled Dot-product Self-Attention as described in equations \ref{eq:1} - \ref{eq:5}. We additionally employ \emph{masking} with a block-diagonal matrix, which enforces locality~\cite{rae-razavi-2020-transformers}. The encoder outputs $\psi_g \in R^D$: a representation of the Google Trend time series enriched with information about which portions of itself are most important to the final exogenous representation for the forecasting task. This information is then fed to the decoder, acting as a prior of the popularity of the product.

\textbf{The transformer decoder} is the component which actually generates the forecast. Unlike the vanilla decoder block of \cite{vaswani2017attention}, we remove the Self-Attention segment, since the input coming from the Feature Fusion Network is a single \textbf{vector} representation and not a sequence. The input is fed to the Multi-Head Cross-Attention attention segment as the query, while $\psi_g$ is projected as the key and query vector. Based on the explanation of the attention mechanism we provided in  Sec. \ref{sec:attention}, this means that our model is trying to understand which part of the exogenous popularity series is most relevant to the multimodal embedding in order to generate a more accurate forecast. The decoder produces a final product embedding $\psi_f \in R^D$, which is a compact representation of four different modalities: $\{\psi_g, \phi_i, \phi_t, \phi_r\}$. Lastly, a fully-connected layer projects $\psi_p$ into $R^{horizon}$ in order to generate the forecasted time series $\{\hat{S}(x, 1),...,\hat{S}(x, horizon)\}$, based on the desired forecast horizon. 

\subsubsection{Training}
Our proposed architecture and all of its components are trained end-to-end using the Mean Squared Error (MSE) loss function, which is minimized using Mini-Batch Gradient Descent via Adafactor \cite{shazeer2018adafactor}. The MSE for a batch of $N$ items is defined as:
\begin{equation}
    MSE(y, \hat{y}) = \frac{1}{N} \sum_{i=1}^N (y-\hat{y})^2
\end{equation}
Minimizing the square of the residuals is naturally a well-known and used method in regression problems, which also transfers to our work. By training our model to minimize this loss, we are trying to generate a point forecast that is as close as possible to the mean of the distribution of the sales process. It is widely known that one of the weak points of least-squares objectives is their sensitivity to outliers, which we try to overcome by performing pattern matching on the exogenous Google Trend signal. In this way, we are anticipating sale peaks by using prior online popularity, which in turn provides a data-driven approach that is both computationally cheap and does not rely on extremely complex modelling concepts.

%% file: src/exp.tex
The experiments start in Sec. \ref{sec:exp:corr} with a preliminary study on how Google trends correlate with the sales. The experimental protocol and implementation details are explained in Sec.~\ref{sec:exp:proto}. In Sec.~\ref{sec:exp:comp} we analyze the first results about how our approach performs against 9 comparative approaches covering the emerging literature of new product sales forecasting. Subsequently in Sec.~\ref{sec:exp:abla}, an ablation study investigates the role of the different modalities we take into account, namely  textual data, image data and the Google trends (see Sec.~\ref{sec:dataset}). The analysis of the performance on the single categories is showed in Sec.~\ref{sec:exp:cat}, while the analysis on different time horizons completes the series of experiments in Sec.~\ref{sec:exp:hor}.    

\subsection{Correlation analysis with Google Trends}\label{sec:exp:corr}
The goal is to check the strength and direction of monotonic association between the sales time series and the Google Trends, motivating their use in our framework. As a preprocessing step, we test the time series for stationarity using the Kwiatkowski-Phillips-Schmidt-Shin (KPSS) test\\~\cite{KPSS}, to make sure that the potential correlations will not be simply due to the temporal component~\cite{aldrich1995correlations}. 34\% of the sales time series are found to be non-stationary and are not considered for this initial analysis.

For each product, we utilize its associated 52-week Google Trends, based on the textual attributes. We calculate the Spearman correlation coefficient against the 12-week sales, using a sliding window protocol with window length $w=12$ and stride of one step. Even though the small sample size does not encourage significance when performing correlation analysis~\cite{de2016comparing}, we wish to investigate the distribution of significant correlations and in particular if they are located on specific periods of the trends. In other words, we are more interested in where the correlations are located across the trends, rather than their values. 

The results give a statistically significant $\rho$ coefficient in 86\% of the total cases. On this selection, the strongest correlations were found to be positive, with 19\% of all coefficients in the range [0.75,1]. The lags that contain the strongest correlations are contained mostly (54\% of the cases) in the range [42,30], i.e., 42-30 weeks before the planned release date.

These findings are quite interesting, since they state that the period which is most correlated to the sales is seven to ten months before the product's release date, which corresponds loosely to the end of the same fashion season from the previous year. This preliminary analysis provides further motivation for the use of the Google Trends and is later confirmed by the cross-attention weights of \approachname in Sec. \ref{sec:exp:interpretability}.

\subsection{Experimental protocol}\label{sec:exp:proto}
On \datasetname{} we define an experimental protocol that simulates how a fast fashion company deals with new products, focusing on two particular moments: i) the \emph{first order setup}, which is when the company orders the first stock of products to be distributed in the shops, usually two months before the starting season; ii) the \emph{release setup}, which is right before the start of the season, and is useful to obtain the best forecast by using all of the exogenous information at hand, so to have a preliminary idea of when to do the stock replenishment. For these two moments we use 28 and 52 timesteps long Google trends, respectively.

As forecast horizon, we consider 6  weeks, as it is the period where no interventions are made by the company, such as reordering or retirements of products (if they perform very poorly). In any case, all models classifiers have been trained assuming a 12-week prediction, and shorter horizons have been taken into account for the evaluation. This procedure maximized the performances of all the approaches. Nonetheless results at different horizons will be shown here as for our approach. To perform the experiments, we divide the data into a training and testing partition, where the testing products are composed of the 497 most recent products. The rest of the dataset (5080 products) is used for training. 

We utilize the \emph{Weighted  Absolute Percentage Error}~\cite{hyndman2008forecasting} as the primary error measure. It expresses the forecasting accuracy as a ratio:
\begin{equation}
    \text{WAPE} = \frac{\sum_{t=1}^{T}|y_t-\hat{y}_t|}{\sum_{t=1}^{T}y_t}
\end{equation}
where $T$ is the forecasting horizon. WAPE is always nonnegative, and a lower value indicates a more accurate model. Even though it is a percentage-based metric, it is not bounded by 100.

For a more articulated understanding of our approach, we compute the \emph{Mean Absolute Error} (MAE), also known as Mean Average Devation (MAD):
\begin{equation}
    \text{MAE} = \frac{\sum_{t=1}^{T}|y_t-\hat{y}_t|}{T}
\end{equation}
MAE describes the mean quantity by which the forecast misses the values on their respective scale.

Forecasting bias~\cite{brown2004smoothing} is another aspect to take into account, measuring systematic over- or underestimation of the forecast w.r.t. the correct value. Even if a slight forecast bias might not have a notable effect on store replenishment, it can lead to over- or under-supply at the central warehouse. To measure the forecasting bias, we adopt the \emph{tracking signal} (TS) measure~\cite{brown2004smoothing,nahmias2009production}:
\begin{equation}
    \text{TS} = \frac{\sum_{t=1}^{T} y_t-\hat{y}_t}{MAE}
\end{equation}
which is basically the signed difference between actual and prediction value, divided by the MAE. The sign of the tracking signal communicates if we have an overestimation (if negative) or an underestimation (if positive). The closer to zero, the more unbiased the forecast. In the literature, a forecasting approach is considered to be consistently biased if the tracking error is above 3.75 or below -3.75~\cite{brown2004smoothing,nahmias2009production}.

Finally, we focus on the capability in providing a forecasting curve which resembles the ground truth, as a way to highlight whether the model has properly captured the actual signal dynamics. To this end, we exploit the Edit distance with Real Penalty (ERP)~\cite{chen2004marriage} which borrows from the classical Edit Distance (ED). ED works on discrete sequences, counting the number of edit operations (insert, delete, replace) that are necessary to transform one series into the other. ERP uses the following algorithm: if the Euclidean distance between prediction $\hat{y}_t$ and $y_t$ is smaller than a penalty $\epsilon$, they are considered equal (d=0) and if not they are considered different (d=1). Summing over differences along the time axis gives the final distance. Since ERP is a dissimilarity, the closer it is to 0 the better.

\subsection{Comparative results}\label{sec:exp:comp}

Comparing \approachname{} with other approaches in the literature requires particular care, since we are the first to exploit Google Trends as exogenous variables to forecast sales for new products. For this reason, together with considering state-of-the-art alternatives in their original form, we adapt them by injecting Google Trends wherever this modification is  natural, for example on models which already do process exogenous data. All the code, including the one for the competitors will be made publicly available, for the sake of fairness. To ease the reading, the name of the approaches will be followed by a square parenthesis indicating the type of information exploited within: T for textual data (category, color, fabric and release date), I for image data, G for google trends. Additionally, the name of the approaches which have been augmented with the Google Trends will be followed by a ``+G''. More in the detail, we consider: 

\textbf{kNN models}. These non-parametric methods methods are proposed in \cite{ekambaram_attention_2020}, and follow a common guideline for fast fashion companies: sales of new products will be similar to older, similar products they have already commercialized~\cite{thomassey2014sales}. The idea is to define a similarity metric between products and then forecast the sales of the new product by averaging the sales of the $k$ most similar products that have sold before. Let $P$ be set of all products and let $d(x_{p_i}, x_{p_j}), \forall x \in P$ be the distance between any two products. We can then obtain the set of $k$ nearest neighbors to a product $K = \{x_1 .. x_k | P, d\}$. We can then estimate the sales of the a product $x_p$ using a weighted average the sales of its neighbors $\sum_{k=1}^{K} \frac{d(x_p, x_k)}{\sum_{k=1}^{K}d(x_p, x_k)}y_k$, where $y$ is the sales time series. The three KNN alternatives proposed in~\cite{ekambaram_attention_2020} are all considered here, which depend on the data they consider to capture the similarity: i) between product attributes (color + category + fabric) \emph{Attribute KNN}; ii) Between product images (\emph{Image KNN}); iii) Between the product attributes \textit{and} images \emph{Attr. + Image KNN}. In our experiments, we use the cosine distance and set $k=11$.

\textbf{Gradient Boosting} (GBoosting) \cite{friedman_2001}. This fundamental technique has been used in time series forecasting either as solitary models~\cite{ henzel2020gradient} and recently as components of more elaborate architectures~\cite{ilic2021explainable}. Gradient Boosting is an ensemble model which aggregate the results from multiple Decision Trees, where we assume Gradient Boosted Trees. Decision Trees are simple, tree-like diagrams for decision making. 
Gradient Boosted Trees build trees one after the other, such that each new tree helps correct the errors made by the previous one. This is done by fitting the trees on the negative of the gradient of a particular loss function (similarly to Backpropagation through SGD in Neural Networks). We use 500 trees and set least squares as the optimization problem. When using this model, the additional features, both exogenous and not, are concatenated together and fed to the model. 

\textbf{Multimodal Encoder-Decoder RNNs}, proposed as most advanced techniques in \cite{ekambaram_attention_2020}. The idea is to perform sequence learning in a two-step process, where an Encoder module takes the available information and produces a learned feature representation of the various modalities. This is then fed to an GRU\cite{cho2014learning} network that acts a Decoder, which autoregressively performs the forecasting. The authors augment their architecture with Bahdanau Attention\cite{bahdanau2016neural}, using the last produced decoder hidden state to learn, at each prediction step, which one of the various modalities provides more important information to the forecast. In particular, we consider the two best performing techniques from the original paper, that is the \emph{Concat Multimodal RNN} (Cat-MM-RNN), which which learns joint embeddings derived by concatenating embeddings of individual input modalities and the \emph{Cross-Attention RNN} (X-Att-RNN), which learns multimodal attention weights and temporal attention weights to create an improved joint embedding. Both these architectures natively accomodate the use of Google Trends, so we feed the trends in the exogenous data module as depicted in \cite{ekambaram_attention_2020}. All neural network models are trained for 200 epochs with a batch size of 128 on an NVIDIA Titan RTX GPU.
\begin{table}[t]
\hspace{-1.45cm}
%\centering
\resizebox{1.2\linewidth}{!}{\begin{tabular}{l|l||c|c|c|c||c|c|c|c}\hline
    \textbf{Method} & \textbf{Input } &  \multicolumn{4}{l||}{\hfil\textbf{6 Weeks, G.Trends: 52 weeks}}&  \multicolumn{4}{l}{\hfil\textbf{6 Weeks, G.Trends: 28 weeks}}\\
    &  &\textbf{WAPE} & \textbf{MAE} & \textbf{TS}& \textbf{ERP}&\textbf{WAPE} & \textbf{MAE} & \textbf{TS}& \textbf{ERP}\\\hline
\emph{Attribute KNN}~\cite{ekambaram_attention_2020} &[T]&59,8&32,7(18;39)&-0,88&0,40&59,8&32,7(18;39)&-0,88&0,40\\ \hline
\emph{ImageKNN}~\cite{ekambaram_attention_2020} &[I]&62,2&34,0(19;42)&-1,09&0,43&62,2&34,0(19;42)&-1,09&0,43\\ \hline
\emph{Attr+Image KNN}~\cite{ekambaram_attention_2020} &[T+I]&61,3&33,5(19;39)&-1,10&0,41&61,3&33,5(19;39)&-1,10&0,41\\ \hline
\emph{GBoosting}~\cite{friedman_2001} &[T+I]&64,1&35,0(21;41)&-1,58&0,43&64,1&35,0(21;41)&-1,58&0,43\\ \hline
\emph{GBoosting+G}~\cite{friedman_2001} &[T+I+G]&63,5&34,7(20;41)&-1,55&0,42&64,3&35,1(21;41)&-1,71&0,43\\ \hline
\emph{Cat-MM-RNN}~\cite{ekambaram_attention_2020} &[T+I]&63,3&34,4(18;44)&-0,67&0,42&63,3&34,4(18;44)&-0,67&0,42\\ \hline
\emph{Cat-MM-RNN+G}~\cite{ekambaram_attention_2020} &[T+I+G]&65,9&35,8(19;45)&-0,41&0,44&64,1&34,8(18;43)&-0,21&0,43\\ \hline
\emph{X-Att-RNN}~\cite{ekambaram_attention_2020} &[T+I]&59,5&32,3(16;39)&-0,32&0,38&59,5&32,3(16;39)&-0,32&0,38\\ \hline
\emph{X-Att-RNN+G}~\cite{ekambaram_attention_2020} &[T+I+G]&59,0&32,1(17;38)&-0,18&0,38&58,7&31,9(16;39)&-0,88&0,38\\ \hline\hline
\textbf{\approachname}&[T+I+G]&55,2&30,2(15;36)&0,41&0,33&56,8&31,0(15;38)&0,90&0,35\\ \hline
\textbf{\approachname**}&[T+I+G]&54,2&29,6(14;35)&0,56&0,33&54,4&29,7(14;36)&0,44&0,31\\ \hline

\end{tabular}}
\vspace{0.15cm}
\caption{Results on \datasetname. \approachname{}** is fed with the  trends indexed by the tags automatically extracted from Fashion IQ. MAE is reported along with its values at the 25-th and 75-th percentiles in parenthesis.}
\label{table:results}
\end{table}

\begin{figure*}[h]
    \hspace{-2cm}
    \includegraphics[width=1.2\linewidth]{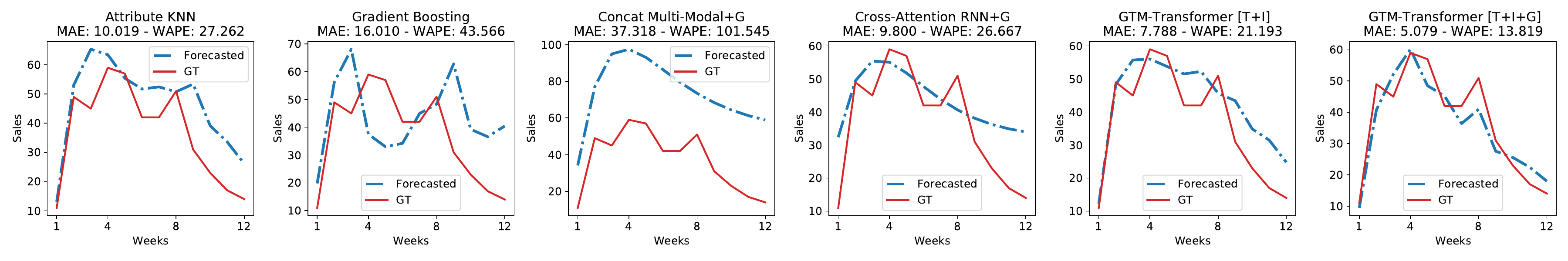}
    
    \hspace{-2cm}
        \includegraphics[width=1.2\linewidth]{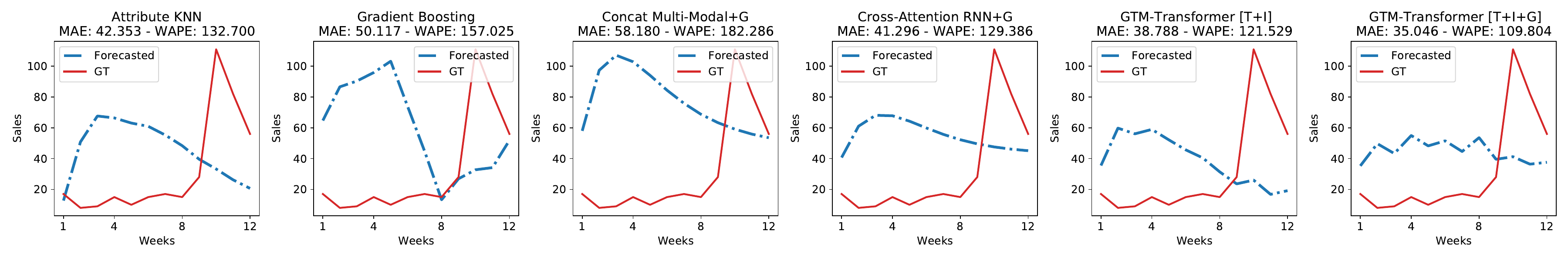}
     \caption{Qualitative results.}
    \label{fig:qualitative}
\end{figure*}

Table~\ref{table:results} reports the results, where the following facts can be pointed out: 
\begin{itemize}[noitemsep]
    \item The use of Google Trends boosts the performance of all the models, except Concat Multimodal RNN , where the Google Trends have been simply concatenated as static data.
    \item Our \approachname gives the best results in both setups (first order and release setup), with the best MAE and WAPE and the second best Tracking Signal, displaying a good balance between over and underestimation. We have the best ERP, which indicates that the shapes of our forecasting curves better resemble the actual sales (as also seen in Fig. ~\ref{fig:qualitative}).
    \item The tracking signal indicates persistent forecasting bias if its value is above (below) 3.75~\cite{brown2004smoothing,nahmias2009production}. Not one of the methods has this problem, including \approachname. This shows that even though the models have gotten much more complex, we are still able to maintain a strong balance between positive and negative errors. \approachname remains balanced even with 28-week Google Trends.
    \item Using shorter Google trends (28-week, Table~\ref{table:results} on the right) gives performances which in general are slightly worse. An explanation for this can be inferred when looking at the attention weights, which are explored in Sec.~\ref{sec:exp:interpretability}
\end{itemize}
To explore the generalization of the model to additional types of visual attributes, we consider the tags from Fashion IQs~\cite{wu2020fashion}: they represent a widely-known approach to describe fashion items for automated retrieval purposes. We apply the attribute extraction code directly to our data, focusing on the ``shape'' attribute, which describes fine-grained aspects of the structure of the product (v-neck, hem, \dots). We discard the other types of attributes, since they consistently overlap with ours (such as the ``fabric'' attribute) or do not fit very well with \datasetname, because in Fashion IQ clothes are worn by models. After the attribute extraction, we download the related Google Trends as described in Sec.~\ref{sec:dataset}. We dub this model in Table~\ref{table:results} as \approachname **. Interestingly, adding complementary information boosts further the model, promoting once again the use of the Google trends.    

Additional insight can be inferred by some qualitative results, showing two 12-week predictions (Fig.~\ref{fig:qualitative}): Attribute KNN gives reasonable estimates, trying to capture the scarce performance of the first 6 weeks portrayed in the second row. Gradient Boosting overestimates both the cases, offering a graphical demonstration of its high tracking signal TS=-1.58 (Table~\ref{table:results}). The RNN-based approaches Concat Multimodal+G, Cross Attention RNN+G seems to have a very regular slope, irrespective of the real structure of the sale signal: this is likely due to the nature of the autoregressive approach, which has learned the general sale curve dynamics and struggles with trajectories which deviate from it. With the \approachname the role of the Google Trends appears to be clear, being capable of giving more structure to the final forecast (above), lowering down the forecasting thus predicting a scarce performance (below).

\subsection{Ablation study}\label{sec:exp:abla}
Ablative results refer to the 6-week forecasting horizon, using the full 52-week Google Trends, and are reported in Tab.~\ref{table:ablative}.   
\begin{table}[h]
%\small
\centering
\resizebox{.6\linewidth}{!}{\begin{tabular}{l||c|c|c|c}\hline
    \textbf{GTM} & \multicolumn{4}{l}{\hfil\textbf{6 Weeks}}\\
    \textbf{ablations} & \textbf{WAPE} & \textbf{MAE (25\%;75\%)} & \textbf{TS}& \textbf{ERP} \\\hline
[I]&56,4&30,8(16;36)&-0,34&0,36\\ \hline
[T]&62,6&34,2(19;43)&-1,42&0,43\\ \hline
[G]&58,2&31,8(17;37)&-0,89&0,38\\ \hline\hline
[I+T]&56,7&30,9(16;38)&-0,32&0,37\\ \hline
[T+G]&56,8&31,0(14;38)&1,63&0,33\\ \hline
[I+G]&55,7&30,4(13;32)&1,45&0,30\\ \hline\hline
\textbf{[T+I+G]}&55,2&30,2(15;36)&0,41&0,33\\ \hline\hline
[AR]&59,6&32,5(14;36)&1,18&0,32\\ \hline
\end{tabular}}
\vspace{0.15cm}
\caption{6 weeks ablative results on \datasetname. MAE is reported with also its values at 25-th and 75-th percentiles in parenthesis.}
\label{table:ablative}
\end{table}

The first ablation is our model without the Google Trends, so removing the encoder module in Fig.~\ref{fig:model_architecture} (row [I+T]). The much higher WAPE  highlights the net role of the exogenous data, and is one of the main results of our study. It is worth noting that the performances are better than all of the approaches using the same kind of information (see Tab.~\ref{table:results}), proving the good design of our architecture.
The two-modality combos text + Google Trends ([T+G]) and image + Google Trends ([I+G]) give WAPE scores both around 57\%, demonstrating that text and images carry complementary information which the complete \approachname is capable of combining and exploiting.
Single modalities ablations instead demonstrate that the image alone [I] has the best performance, and this obviously states that it is the appearance of the product which allows for the most discrimination. Surprisingly, Google Trends [G] alone gives the second best results, while text attributes [T] alone gives the worst results, indicating once again the net value of this exogenous signal.   

Finally, the \emph{[AR]} row indicates the complete model, but in its autoregressive version: the performance is 4.4\% worse than our \approachname, showing the benefit of the non-autoregressive design.

\subsection{Single category analysis}\label{sec:exp:cat}
\begin{figure}[h]
    \centering
    \includegraphics[width=0.8\linewidth]{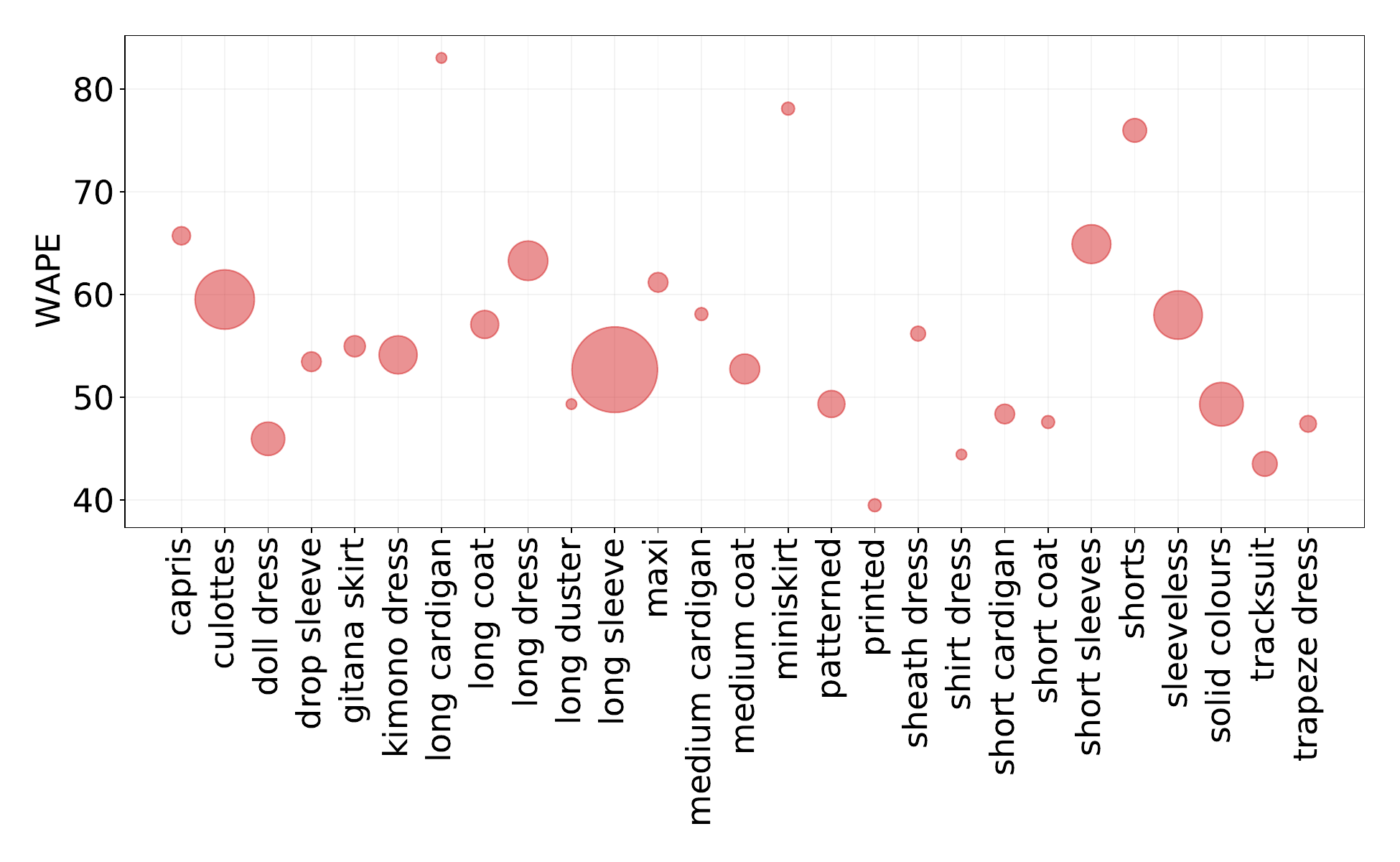}
     \caption{Forecasting performance for different categories.}
    \label{fig:category_results}
\end{figure}

Is interesting to check how \approachname performs on different categories. Figure~\ref{fig:category_results} contains the separate WAPEs, where the marker size represents the cardinality of the category (Fig.~\ref{fig:cardinalities}a). The results confirm the fact that performances are more stable for categories with a large number of products such as "Long sleeve" or "Culottes", as the amount of data available for training over these products is larger.

\subsection{Varying the forecasting horizon}\label{sec:exp:hor}
\begin{figure}[h]
    \centering
    \includegraphics[width=0.8\linewidth]{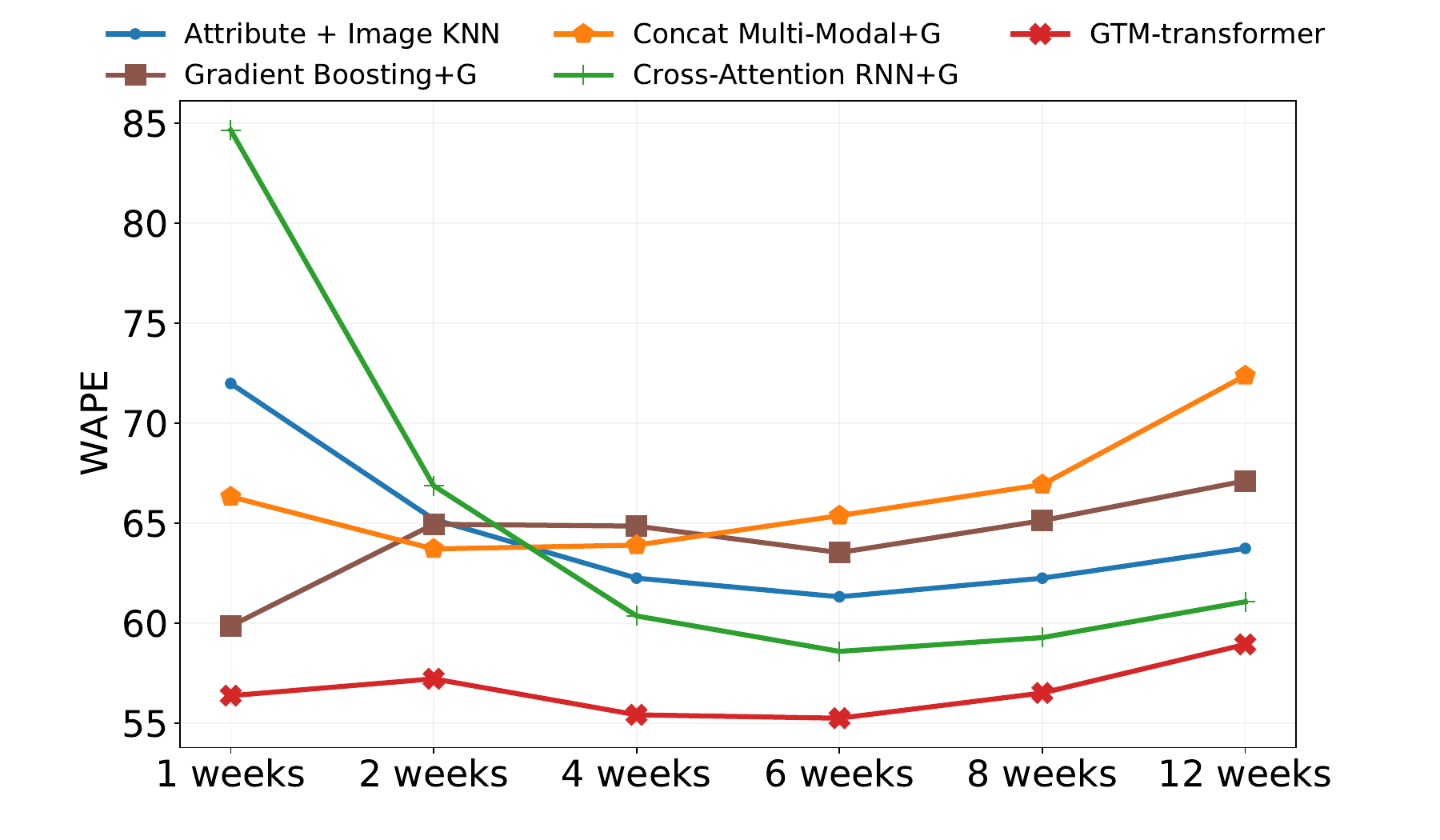}
    \caption{Forecasting performance for different forecasting horizons.}
    \label{fig:hor_results}
\end{figure}

In this section we demonstrate the effect of the forecasting horizon on the performance. Figure~\ref{fig:hor_results} contains the WAPE for 1, 2, 4, 6, 8 and 12 week forecasts.  \approachname remains the best performing approach for all horizons, on pair at 2 weeks with Cross-Attention RNN+G. Most of the slopes show a minimum error at 6 weeks, except the Gradient Boosting which shows the second best performance at 1 week. The first 6 weeks performance varies greatly, with Attribute + Image KNN performing the worst. After 6 weeks, all the approaches have a decrease in the performance, which is natural, since the sale signal becomes more dependent on external choices (replenishments, discounts) we are not modeling here.

\subsection{Model interpretability: unveiling the Google Trends}\label{sec:exp:interpretability}
To understand the role of Google Trends in \approachname we exploit the interpretability of the Attention mechanism. To this sake we calculate where in the Google Trend the decoder assigns the highest Cross-Attention weight for the testing set, to find if there are any systematical tendencies as to where the model looks at when making the prediction. A windown binning approach is used, similarly to the correlation analysis in Sec. \ref{sec:exp:corr}, and we count the number of times that the lag with the highest weight is located in that particular window. Table \ref{table:attn_lag} contains the results, where it can be seen that the initial period of the Google Trend seems to be the most crucial, as also hinted initially by our correlation analysis.

\begin{table}[h]%{r}%{2.5cm}
\footnotesize
\centering
\resizebox{.7\linewidth}{!}{\begin{tabular}{l||c|c|c|c|c}\hline
    \textbf{Lag} & \emph{52 $-$ 40} & \emph{42 $-$ 30} & \emph{32 $-$ 20} & \emph{22 $-$ 10} & \emph{12 $-$ 0} \\\hline
    \textbf{\#Highest W} & 145 & 231 & 42 & 46 & 33 \\ \hline
\end{tabular}}
\vspace{0.15cm}
\caption{Points of the Google Trends time series with the highest Cross-attention weights.}
\label{table:attn_lag}
\end{table}

\subsection{A very practical use of our model: the \emph{first-order} problem}\label{sec:exp:fo}
Accurate new product forecasting is highly desirable for many reasons, as explained in the introduction: understand tendency in the sales, deciding when to replenish the warehouses, and how many products per reference to buy before the season starts. This is known as the \emph{first-order} problem~\cite{donohue2000efficient}, and it can be accurately simulated with the real data of \datasetname.  The goal is to order a number of products that matches the sum of future sales until the sixth week, without exceeding or underestimating. During the first six weeks then, sales will help with more predictive power in suggesting how to behave with the remaining weeks, for example deciding whether to order again or not.   

A general protocol to deal with the first order problem is to consider the sum of the sold products of the same period in the previous correspondent season, adding a percentage which mirrors the expected growth, and make the first order. In our case, the policy adopted by the company is to increase the orders for a product of a particular category, color and fabric by 60\% of the previous average sum of sold products in the first six weeks for those attributes. We call this the 60\% policy. For example, if we want to do the first order for SS19 season of a new white cotton cardigan, we take the average of sold white cotton cardigans of the previous SS18 and add the 60\%.

To compute the first order error, we simply calculate the integral of the forecasting and ground truth curves for the first 6 weeks and compare them with each other, for each considered approach, including the 60\% policy. To evaluate the performance, we compute the mean of all the absolute errors over all products. This tells us by how much, on average, the model is mistaken about the total sold amount and therefore the quantity of the first order. To show the real impact of such a problem, in Table~\ref{table:firstOrder} we report also the monetary discrepancy in US dollars, assuming that each reference has a cost of \$25 (the average cost of a fast fashion product). In a market of around 12M dollars, the 60\% policy is clearly ineffective, and all the forecasting approaches lower the discrepancy considerably, with \approachname lowering it the most. 
\begin{table}[h]
%\footnotesize
\small
\centering
\begin{tabular}{l|c|c}\hline
    \textbf{Method} & \multicolumn{2}{l}{\hfil\textbf{6 Weeks}}\\
    &  \textbf{MAE} & \textbf{US \$ discr. \textbf{$\downarrow$}} \\\hline
\emph{60\% Policy} & 313,6 & 3.920.125 \$\\ \hline
\emph{Attribute KNN}~\cite{ekambaram_attention_2020}&271,0&3.366.863 \$\\ \hline
\emph{ImageKNN}~\cite{ekambaram_attention_2020}&279,7&3.475.242 \$\\ \hline
\emph{Attribute + Image KNN}~\cite{ekambaram_attention_2020}&271,9&3.378.441 \$\\ \hline
\emph{Gradient Boosting+G}~\cite{ilic2021explainable}&297,2&3.692.453 \$\\ \hline
\emph{Concat Multimodal KNN+G}~\cite{ekambaram_attention_2020}&359,7&4.495.977 \$\\ \hline
\emph{Cross-Attention RNN+G}~\cite{ekambaram_attention_2020}&271,5&3.393.695 \$\\ \hline\hline
\textbf{\approachname}&\textbf{262,3}&\textbf{3.236.753} \$\\ \hline

\end{tabular}
\vspace{0.15cm}
\caption{First-order results on \datasetname. }
\label{table:firstOrder}
\end{table}

%% file: src/conc.tex
In this paper we tackle the problem of new fashion product sales forecasting, which is a challenge that requires alternative solutions powered by machine learning and computer vision. In this scenario, we show that Google Trends are beneficial exogenous data and help augment the model's reasoning by adding a popularity prior for the items. All of this sout{was} \textbf{is} possible thanks to a multimodal framework based on the Transformer, made non-autoregressive in order to deal with the complex dynamics which sales data exhibit, by effectively ingesting the Google Trends data. Additionally and thanks to the collaboration of Nunalie, a genuine dataset coming from the company's recent past sales has been proposed and made publicly available, equipped with ground truth sales signals and data from the image and text domain\textbf{s}. Multiple directions can be considered for future work, starting from the use of a more "representative" popularity signal (data-centric aspect), to probabilistic modelling of the complex sales distributions of this practical and challenging problem. 

\paragraph{Acknowledgements.} This work has been partially supported by the project of the Italian Ministry of Education, Universities and Research (MIUR) "Dipartimenti di Eccellenza 2018-2022". We would like to thank Nunalie for their availability and for allowing us to publish \datasetname. 

\clearpage